\pdfoutput=1

\documentclass[11pt]{article}

\usepackage[final]{ACL2023}

\usepackage{times}
\usepackage{latexsym}
\usepackage{graphicx}
\usepackage{makecell}
\usepackage{float}
\usepackage{multirow}
\usepackage{natbib}
\usepackage{colortbl}
\usepackage{xcolor}
\usepackage{placeins} 
\usepackage{listings}
\definecolor{lightred}{rgb}{1.0, 0.7, 0.7}
\definecolor{lightgreen}{rgb}{0.7, 1.0, 0.7}

\usepackage[T1]{fontenc}

\usepackage[utf8]{inputenc}

\usepackage{microtype}

\usepackage{inconsolata}

\title{Better To Ask in English? Evaluating Factual Accuracy of Multilingual LLMs in English and Low-Resource Languages}

\author{
 \textbf{Pritika Rohera\textsuperscript{1,3}},
 \textbf{Chaitrali Ginimav\textsuperscript{1,3}},
 \textbf{Gayatri Sawant\textsuperscript{1,3}}, and
 \textbf{Raviraj Joshi\textsuperscript{2,3}}
\\
 Pune Institute of Computer Technology\textsuperscript {1} \\
 Indian Institute of Technology Madras\textsuperscript {2} \\
 L3Cube Labs, Pune\textsuperscript{3}
}

\begin{document}
\maketitle
\begin{abstract}
Multilingual Large Language Models (LLMs) have demonstrated significant effectiveness across various languages, particularly in high-resource languages such as English. However, their performance in terms of factual accuracy across other low-resource languages, especially Indic languages, remains an area of investigation. In this study, we assess the factual accuracy of LLMs — GPT-4o, Gemma-2-9B, Gemma-2-2B, and Llama-3.1-8B - by comparing their performance in English and Indic languages using the IndicQuest dataset, which contains question-answer pairs in English and 19 Indic languages. By asking the same questions in English and their respective Indic translations, we analyze whether the models are more reliable for regional context questions in Indic languages or when operating in English. Our findings reveal that LLMs often perform better in English, even for questions rooted in Indic contexts. Notably, we observe a higher tendency for hallucination in responses generated in low-resource Indic languages, highlighting challenges in the multilingual understanding capabilities of current LLMs.
\end{abstract}

\section{Introduction}

The rapid advancement of Large Language Models (LLMs) has significantly transformed Natural Language Processing (NLP), enabling remarkable performance across diverse tasks. However, while these models excel in high-resource languages like English, they face notable challenges when applied to low-resource Indic languages \cite{singh-etal-2024-indicgenbench}. Due to the limited availability of digitized resources and training data for these languages, LLMs often struggle to generate factually accurate responses \cite{verma2025milumultitaskindiclanguage}. This issue is particularly critical in domain-specific regional contexts, such as history, geography, politics, and economics, where LLMs frequently produce factually incorrect or "hallucinated" responses in Indic languages. Inaccuracies for Indic languages commonly appear as distorted or incorrect representations of known facts, or even fabricated information \cite{liu2024enhancinglargelanguagemodels}.

\begin{figure}[H]
    \centering
    \includegraphics[width=\columnwidth]{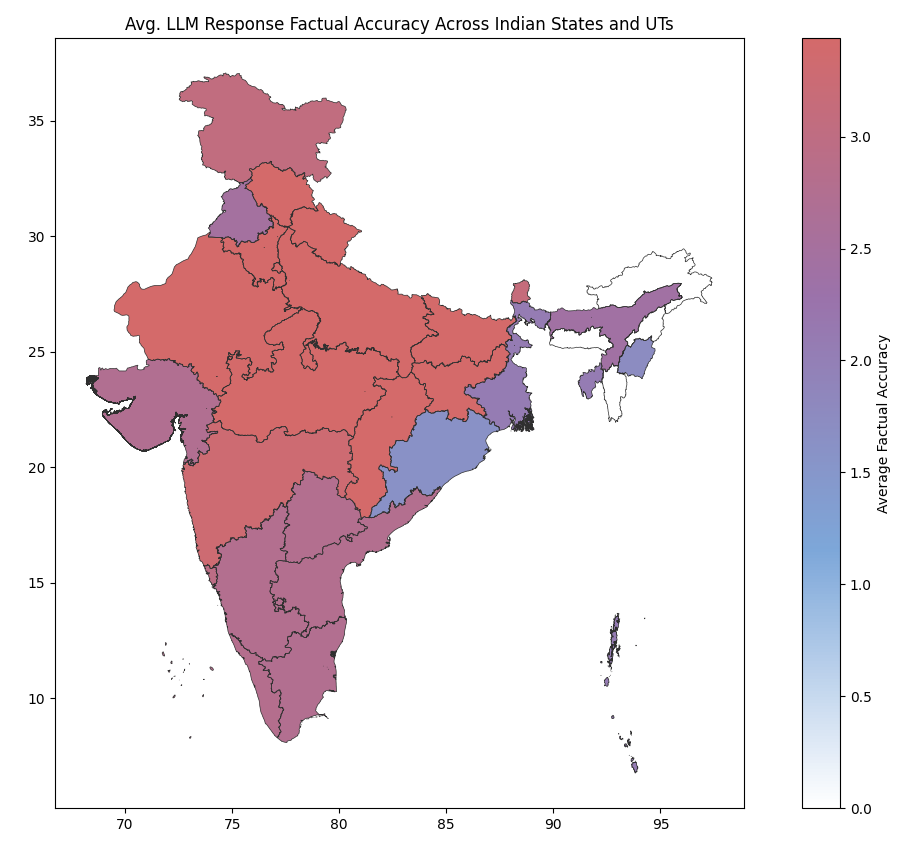}
    \caption{Heatmap showing the average factual accuracy of LLM-generated responses across Indian states and union territories, based on their primary language. Regions not represented in the dataset are marked in white. See Appendix Figure \ref{fig:factual_accuracy_heatmap_appendix} for full language-region mappings and scoring details.}
    \label{fig:factual_accuracy_heatmapl}
\end{figure}

These inaccuracies are particularly concerning for Indic languages, making it crucial to assess the performance of multilingual LLMs in these settings. Understanding their ability to handle domain-specific regional questions in low-resource Indic languages can provide valuable insights into the models' limitations and potential areas for improvement. 

This paper evaluates the response capabilities of multilingual LLMs in English and 19 Indic languages, namely Assamese, Bengali, Dogri, Gujarati, Hindi, Kannada, Konkani, Maithili, Malayalam, Marathi, Meitei (Manipuri), Nepali, Odia, Punjabi, Sanskrit, Sindhi, Tamil, Telugu, and Urdu, with a focus on factual accuracy. By examining their answers to identical factual questions in English and low-resource Indic languages, we aim to highlight performance disparities. Our analysis utilizes the IndicQuest dataset \cite{rohera2024l3cubeindicquestbenchmarkquestinganswering}, which features factual questions across various domains in English and selected Indic languages. We evaluate four prominent LLMs, namely GPT-4o, Gemma-2-9B, Gemma-2-2B, and Llama-3.1-8B.

\begin{figure}[]
    \centering
    \includegraphics[width=\columnwidth]{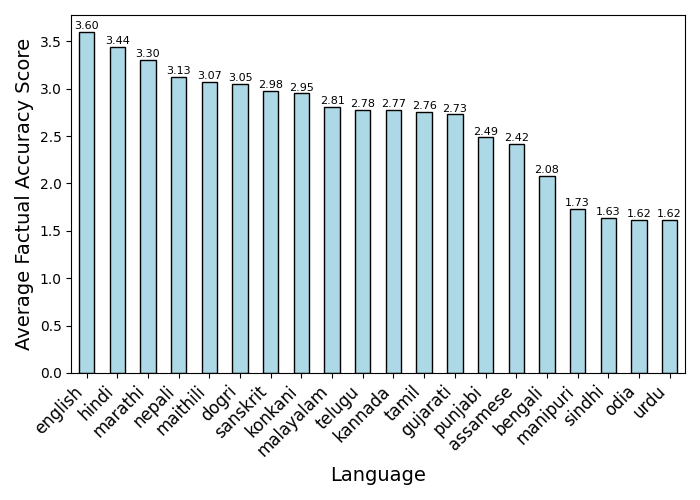}
    \caption{Average factual accuracy scores for all 20 languages in the dataset. The scores are computed by aggregating factual accuracy ratings of LLM-generated responses across all evaluated models for each language.}
    \label{fig:factual_accuracy_by_language}
\end{figure}

Through detailed, comprehensive performance analysis, we address the following key questions:
\begin{enumerate}
    \item \textbf{How well do multilingual LLMs maintain factual accuracy across different languages, particularly in region-specific contexts?} While we ideally expect these models to perform better on regional Indian context questions when prompted in Indic languages, our observations reveal that this expectation does not hold true.
    \item \textbf{How does the size of the model impact factual accuracy?} While larger models generally exhibit higher factual accuracy, our analysis reveals that this improvement is not uniformly distributed across languages, with low-resource Indic languages showing minimal gains compared to English.
    \item \textbf{How does the choice of domain affect the factual accuracy of responses?} The choice of domain significantly impacts factual accuracy, with English consistently outperforming Indic languages across all domains. Notably, Economics and Literature show relatively good performance across languages, while History and Geography exhibit the lowest accuracy.

    \item \textbf{Is there a correlation between language consistency and factual accuracy?} We observe a strong correlation between language consistency and factual accuracy, with higher consistency scores typically aligning with better accuracy.
\end{enumerate}

\begin{table*}
\centering
\small
\setlength{\tabcolsep}{3pt}
\renewcommand{\arraystretch}{1.25}
\resizebox{\textwidth}{!}{%
\begin{tabular}{|l|c|*{20}{c|}}
\hline
\multirow{2}{*}{\textbf{Metric}} & \multirow{2}{*}{\textbf{Model}} & \multicolumn{20}{c|}{\textbf{Language Scores}} \\
\cline{3-22}
 & & As & Be & Do & En & Gu & Hi & Ka & Ko & Mi & Ml & Mn & Mr & Ne & Od & Pu & Sa & Si & Ta & Te & Ur \\
\hline
\multirow{4}{*}{Factual Accuracy} 
 & GPT-4o & - & - & - & \cellcolor{lightgreen}4.24 & - & \cellcolor{lightgreen}4.32 & - & - & - & - & - & \cellcolor{lightgreen}4.11 & - & - & - & - & - & - & - & - \\
 & Gemma-2-9b & \cellcolor{lightgreen}3.16 & \cellcolor{lightgreen}2.69 & \cellcolor{lightgreen}3.50 & \cellcolor{lightgreen}3.69 & \cellcolor{lightgreen}3.42 & \cellcolor{lightgreen}3.45 & \cellcolor{lightgreen}3.54 & \cellcolor{lightgreen}3.38 & \cellcolor{lightgreen}3.49 & \cellcolor{lightgreen}3.51 & \cellcolor{lightred}1.66 & \cellcolor{lightgreen}3.42 & \cellcolor{lightgreen}3.66 & \cellcolor{lightred}1.94 & \cellcolor{lightgreen}3.08 & \cellcolor{lightgreen}3.57 & \cellcolor{lightred}2.12 & \cellcolor{lightgreen}3.43 & \cellcolor{lightgreen}3.54 & \cellcolor{lightred}1.94 \\
 & Llama-3.1-8b & \cellcolor{lightred}2.37 & \cellcolor{lightred}1.95 & \cellcolor{lightgreen}3.03 & \cellcolor{lightgreen}3.35 & \cellcolor{lightgreen}2.83 & \cellcolor{lightgreen}3.28 & \cellcolor{lightgreen}2.67 & \cellcolor{lightgreen}2.93 & \cellcolor{lightgreen}3.08 & \cellcolor{lightgreen}2.75 & \cellcolor{lightgreen}2.52 & \cellcolor{lightgreen}3.03 & \cellcolor{lightgreen}3.02 & \cellcolor{lightred}1.78 & \cellcolor{lightgreen}2.71 & \cellcolor{lightgreen}2.67 & \cellcolor{lightred}1.49 & \cellcolor{lightgreen}2.55 & \cellcolor{lightgreen}2.86 & \cellcolor{lightred}1.78 \\
 & Gemma-2-2b & \cellcolor{lightred}1.72 & \cellcolor{lightred}1.59 & \cellcolor{lightgreen}2.62 & \cellcolor{lightgreen}3.11 & \cellcolor{lightred}1.94 & \cellcolor{lightgreen}2.72 & \cellcolor{lightred}2.10 & \cellcolor{lightgreen}2.53 & \cellcolor{lightgreen}2.65 & \cellcolor{lightred}2.16 & \cellcolor{lightred}1.02 & \cellcolor{lightgreen}2.65 & \cellcolor{lightgreen}2.70 & \cellcolor{lightred}1.13 & \cellcolor{lightred}1.66 & \cellcolor{lightgreen}2.69 & \cellcolor{lightred}1.30 & \cellcolor{lightred}2.30 & \cellcolor{lightred}1.93 & \cellcolor{lightred}1.13 \\
\hline
\multirow{4}{*}{Relevance} 
 & GPT-4o & - & - & - & \cellcolor{lightgreen}4.87 & - & \cellcolor{lightgreen}4.90 & - & - & - & - & - & \cellcolor{lightgreen}4.74 & - & - & - & - & - & - & - & - \\
 & Gemma-2-9b & \cellcolor{lightgreen}4.39 & \cellcolor{lightgreen}3.85 & \cellcolor{lightgreen}4.52 & \cellcolor{lightgreen}4.75 & \cellcolor{lightgreen}4.53 & \cellcolor{lightgreen}4.75 & \cellcolor{lightgreen}4.39 & \cellcolor{lightgreen}4.34 & \cellcolor{lightgreen}4.53 & \cellcolor{lightgreen}4.44 & \cellcolor{lightred}1.94 & \cellcolor{lightgreen}4.60 & \cellcolor{lightgreen}4.68 & \cellcolor{lightgreen}2.81 & \cellcolor{lightgreen}4.47 & \cellcolor{lightgreen}4.63 & \cellcolor{lightgreen}3.19 & \cellcolor{lightgreen}4.45 & \cellcolor{lightgreen}4.37 & \cellcolor{lightgreen}2.81 \\
 & Llama-3.1-8b & \cellcolor{lightgreen}3.70 & \cellcolor{lightgreen}3.09 & \cellcolor{lightgreen}4.26 & \cellcolor{lightgreen}4.48 & \cellcolor{lightgreen}4.21 & \cellcolor{lightgreen}4.54 & \cellcolor{lightgreen}3.88 & \cellcolor{lightgreen}4.04 & \cellcolor{lightgreen}4.37 & \cellcolor{lightgreen}3.75 & \cellcolor{lightgreen}3.25 & \cellcolor{lightgreen}4.31 & \cellcolor{lightgreen}4.28 & \cellcolor{lightgreen}2.87 & \cellcolor{lightgreen}4.18 & \cellcolor{lightgreen}4.00 & \cellcolor{lightgreen}2.81 & \cellcolor{lightgreen}3.40 & \cellcolor{lightgreen}3.83 & \cellcolor{lightgreen}2.87 \\
 & Gemma-2-2b & \cellcolor{lightgreen}2.82 & \cellcolor{lightgreen}2.65 & \cellcolor{lightgreen}3.70 & \cellcolor{lightgreen}4.52 & \cellcolor{lightgreen}3.31 & \cellcolor{lightgreen}4.23 & \cellcolor{lightgreen}3.03 & \cellcolor{lightgreen}3.51 & \cellcolor{lightgreen}3.87 & \cellcolor{lightgreen}3.02 & \cellcolor{lightred}1.07 & \cellcolor{lightgreen}3.87 & \cellcolor{lightgreen}3.84 & \cellcolor{lightred}1.22 & \cellcolor{lightgreen}2.92 & \cellcolor{lightgreen}3.70 & \cellcolor{lightred}1.61 & \cellcolor{lightgreen}3.41 & \cellcolor{lightgreen}2.93 & \cellcolor{lightred}1.22 \\
\hline
\multirow{4}{*}{Clarity} 
 & GPT-4o & - & - & - & \cellcolor{lightgreen}4.74 & - & \cellcolor{lightgreen}4.55 & - & - & - & - & - & \cellcolor{lightgreen}4.20 & - & - & - & - & - & - & - & - \\
 & Gemma-2-9b & \cellcolor{lightgreen}4.10 & \cellcolor{lightgreen}3.23 & \cellcolor{lightgreen}4.29 & \cellcolor{lightgreen}4.71 & \cellcolor{lightgreen}4.13 & \cellcolor{lightgreen}4.52 & \cellcolor{lightgreen}3.93 & \cellcolor{lightgreen}3.95 & \cellcolor{lightgreen}4.38 & \cellcolor{lightgreen}3.99 & \cellcolor{lightred}1.93 & \cellcolor{lightgreen}4.17 & \cellcolor{lightgreen}4.33 & \cellcolor{lightgreen}2.77 & \cellcolor{lightgreen}4.21 & \cellcolor{lightgreen}4.29 & \cellcolor{lightgreen}3.34 & \cellcolor{lightgreen}3.97 & \cellcolor{lightgreen}3.89 & \cellcolor{lightgreen}2.77 \\
 & Llama-3.1-8b & \cellcolor{lightgreen}3.55 & \cellcolor{lightgreen}2.58 & \cellcolor{lightgreen}4.15 & \cellcolor{lightgreen}4.60 & \cellcolor{lightgreen}3.98 & \cellcolor{lightgreen}4.46 & \cellcolor{lightgreen}3.68 & \cellcolor{lightgreen}3.97 & \cellcolor{lightgreen}4.28 & \cellcolor{lightgreen}3.54 & \cellcolor{lightgreen}2.90 & \cellcolor{lightgreen}4.07 & \cellcolor{lightgreen}3.96 & \cellcolor{lightgreen}3.08 & \cellcolor{lightgreen}4.03 & \cellcolor{lightgreen}3.75 & \cellcolor{lightgreen}3.00 & \cellcolor{lightgreen}3.43 & \cellcolor{lightgreen}3.53 & \cellcolor{lightgreen}3.08 \\
 & Gemma-2-2b & \cellcolor{lightgreen}2.72 & \cellcolor{lightred}2.25 & \cellcolor{lightgreen}4.07 & \cellcolor{lightgreen}4.62 & \cellcolor{lightgreen}2.96 & \cellcolor{lightgreen}4.16 & \cellcolor{lightgreen}2.62 & \cellcolor{lightgreen}3.26 & \cellcolor{lightgreen}4.02 & \cellcolor{lightgreen}2.77 & \cellcolor{lightgreen}3.94 & \cellcolor{lightgreen}3.54 & \cellcolor{lightgreen}3.64 & \cellcolor{lightred}2.39 & \cellcolor{lightgreen}2.94 & \cellcolor{lightgreen}3.69 & \cellcolor{lightgreen}2.94 & \cellcolor{lightgreen}3.08 & \cellcolor{lightgreen}2.66 & \cellcolor{lightred}2.39 \\
\hline
\multirow{4}{*}{Language Consistency} 
 & GPT-4o & - & - & - & \cellcolor{lightgreen}5.00 & - & \cellcolor{lightgreen}4.99 & - & - & - & - & - & \cellcolor{lightgreen}5.00 & - & - & - & - & - & - & - & - \\
 & Gemma-2-9b & \cellcolor{lightgreen}4.97 & \cellcolor{lightgreen}4.75 & \cellcolor{lightgreen}4.88 & \cellcolor{lightgreen}5.00 & \cellcolor{lightgreen}5.00 & \cellcolor{lightgreen}5.00 & \cellcolor{lightgreen}4.96 & \cellcolor{lightgreen}4.98 & \cellcolor{lightgreen}4.96 & \cellcolor{lightgreen}4.98 & \cellcolor{lightgreen}4.16 & \cellcolor{lightgreen}5.00 & \cellcolor{lightgreen}5.00 & \cellcolor{lightgreen}4.61 & \cellcolor{lightgreen}4.85 & \cellcolor{lightgreen}4.92 & \cellcolor{lightgreen}4.15 & \cellcolor{lightgreen}4.98 & \cellcolor{lightgreen}4.98 & \cellcolor{lightgreen}4.61 \\
 & Llama-3.1-8b & \cellcolor{lightgreen}4.96 & \cellcolor{lightgreen}4.52 & \cellcolor{lightgreen}5.00 & \cellcolor{lightgreen}5.00 & \cellcolor{lightgreen}4.98 & \cellcolor{lightgreen}5.00 & \cellcolor{lightgreen}4.98 & \cellcolor{lightgreen}4.98 & \cellcolor{lightgreen}5.00 & \cellcolor{lightgreen}4.96 & \cellcolor{lightgreen}5.00 & \cellcolor{lightgreen}4.98 & \cellcolor{lightgreen}4.98 & \cellcolor{lightgreen}4.92 & \cellcolor{lightgreen}4.94 & \cellcolor{lightgreen}4.98 & \cellcolor{lightgreen}4.94 & \cellcolor{lightgreen}5.00 & \cellcolor{lightgreen}4.98 & \cellcolor{lightgreen}4.92 \\
 & Gemma-2-2b & \cellcolor{lightgreen}4.78 & \cellcolor{lightgreen}4.53 & \cellcolor{lightgreen}3.85 & \cellcolor{lightgreen}5.00 & \cellcolor{lightgreen}4.68 & \cellcolor{lightgreen}4.69 & \cellcolor{lightgreen}4.88 & \cellcolor{lightgreen}4.55 & \cellcolor{lightgreen}4.06 & \cellcolor{lightgreen}4.65 & \cellcolor{lightred}1.02 & \cellcolor{lightgreen}4.74 & \cellcolor{lightgreen}4.48 & \cellcolor{lightred}2.17 & \cellcolor{lightgreen}4.02 & \cellcolor{lightgreen}3.96 & \cellcolor{lightred}2.37 & \cellcolor{lightgreen}4.65 & \cellcolor{lightgreen}4.58 & \cellcolor{lightred}2.17 \\
\hline
\multirow{4}{*}{Conciseness} 
 & GPT-4o & - & - & - & \cellcolor{lightgreen}4.23 & - & \cellcolor{lightgreen}4.07 & - & - & - & - & - & \cellcolor{lightgreen}3.92 & - & - & - & - & - & - & - & - \\
 & Gemma-2-9b & \cellcolor{lightgreen}3.88 & \cellcolor{lightgreen}3.29 & \cellcolor{lightgreen}3.98 & \cellcolor{lightgreen}4.22 & \cellcolor{lightgreen}3.97 & \cellcolor{lightgreen}4.26 & \cellcolor{lightgreen}3.78 & \cellcolor{lightgreen}3.72 & \cellcolor{lightgreen}4.13 & \cellcolor{lightgreen}3.86 & \cellcolor{lightred}1.86 & \cellcolor{lightgreen}4.00 & \cellcolor{lightgreen}4.10 & \cellcolor{lightgreen}2.59 & \cellcolor{lightgreen}4.04 & \cellcolor{lightgreen}4.05 & \cellcolor{lightgreen}3.04 & \cellcolor{lightgreen}3.82 & \cellcolor{lightgreen}3.78 & \cellcolor{lightgreen}2.59 \\
 & Llama-3.1-8b & \cellcolor{lightgreen}3.49 & \cellcolor{lightgreen}2.90 & \cellcolor{lightgreen}4.02 & \cellcolor{lightgreen}4.12 & \cellcolor{lightgreen}3.88 & \cellcolor{lightgreen}4.38 & \cellcolor{lightgreen}3.70 & \cellcolor{lightgreen}3.94 & \cellcolor{lightgreen}4.19 & \cellcolor{lightgreen}3.46 & \cellcolor{lightgreen}2.89 & \cellcolor{lightgreen}4.03 & \cellcolor{lightgreen}3.84 & \cellcolor{lightgreen}3.13 & \cellcolor{lightgreen}3.91 & \cellcolor{lightgreen}3.76 & \cellcolor{lightgreen}2.98 & \cellcolor{lightgreen}3.50 & \cellcolor{lightgreen}3.56 & \cellcolor{lightgreen}3.13 \\
 & Gemma-2-2b & \cellcolor{lightgreen}2.53 & \cellcolor{lightred}2.19 & \cellcolor{lightgreen}3.51 & \cellcolor{lightgreen}3.92 & \cellcolor{lightgreen}2.67 & \cellcolor{lightgreen}3.78 & \cellcolor{lightred}2.41 & \cellcolor{lightgreen}2.81 & \cellcolor{lightgreen}3.45 & \cellcolor{lightgreen}2.52 & \cellcolor{lightgreen}3.21 & \cellcolor{lightgreen}3.22 & \cellcolor{lightgreen}3.23 & \cellcolor{lightred}1.67 & \cellcolor{lightgreen}2.69 & \cellcolor{lightgreen}3.23 & \cellcolor{lightred}2.12 & \cellcolor{lightgreen}2.77 & \cellcolor{lightred}2.43 & \cellcolor{lightred}1.67 \\
 \hline
 \multirow{4}{*}{Overall} 
 & GPT-4o & - & - & - & \cellcolor{lightgreen}4.45 & - & \cellcolor{lightgreen}4.49 & - & - & - & - & - & \cellcolor{lightgreen}4.27 & - & - & - & - & - & - & - & - \\
 & Gemma-2-9b & \cellcolor{lightgreen}3.78 & \cellcolor{lightgreen}3.28 & \cellcolor{lightgreen}4.03 & \cellcolor{lightgreen}4.17 & \cellcolor{lightgreen}3.93 & \cellcolor{lightgreen}4.11 & \cellcolor{lightgreen}3.94 & \cellcolor{lightgreen}3.83 & \cellcolor{lightgreen}4.05 & \cellcolor{lightgreen}3.95 & \cellcolor{lightred}2.26 & \cellcolor{lightgreen}3.98 & \cellcolor{lightgreen}4.14 & \cellcolor{lightgreen}2.70 & \cellcolor{lightgreen}3.79 & \cellcolor{lightgreen}4.06 & \cellcolor{lightgreen}2.86 & \cellcolor{lightgreen}3.90 & \cellcolor{lightgreen}3.92 & \cellcolor{lightgreen}2.70 \\
 & Llama-3.1-8b & \cellcolor{lightgreen}3.25 & \cellcolor{lightgreen}2.72 & \cellcolor{lightgreen}3.78 & \cellcolor{lightgreen}3.98 & \cellcolor{lightgreen}3.62 & \cellcolor{lightgreen}4.01 & \cellcolor{lightgreen}3.44 & \cellcolor{lightgreen}3.63 & \cellcolor{lightgreen}3.83 & \cellcolor{lightgreen}3.44 & \cellcolor{lightgreen}3.17 & \cellcolor{lightgreen}3.78 & \cellcolor{lightgreen}3.71 & \cellcolor{lightgreen}2.79 & \cellcolor{lightgreen}3.58 & \cellcolor{lightgreen}3.49 & \cellcolor{lightgreen}2.58 & \cellcolor{lightgreen}3.29 & \cellcolor{lightgreen}3.52 & \cellcolor{lightgreen}2.79 \\
 & Gemma-2-2b & \cellcolor{lightgreen}2.58 & \cellcolor{lightred}2.38 & \cellcolor{lightgreen}3.29 & \cellcolor{lightgreen}3.81 & \cellcolor{lightgreen}2.79 & \cellcolor{lightgreen}3.55 & \cellcolor{lightgreen}2.80 & \cellcolor{lightgreen}3.10 & \cellcolor{lightgreen}3.34 & \cellcolor{lightgreen}2.82 & \cellcolor{lightred}1.83 & \cellcolor{lightgreen}3.28 & \cellcolor{lightgreen}3.32 & \cellcolor{lightred}1.66 & \cellcolor{lightgreen}2.55 & \cellcolor{lightgreen}3.24 & \cellcolor{lightred}1.95 & \cellcolor{lightgreen}3.00 & \cellcolor{lightgreen}2.70 & \cellcolor{lightred}1.66 \\

\hline
\end{tabular}
}
\caption{Metric scores across languages for different models. Scores below 2.5 highlighted in red (poor) and scores above 2.5 in green (better) for quick visual interpretation. Abbreviations for languages are as follows: As (Assamese), Be (Bengali), Do (Dogri), En (English), Gu (Gujarati), Hi (Hindi), Ka (Kannada), Ko (Konkani), Mi (Maithili), Ml (Malayalam), Mn (Manipuri), Mr (Marathi), Ne (Nepali), Od (Odia), Pu (Punjabi), Sa (Sanskrit), Si (Sindhi), Ta (Tamil), Te (Telugu), Ur (Urdu)}
\label{tab:metrics_scores}
\end{table*}

\section{Related Works}
Recent advances in multilingual large language models (MLLMs) have significantly expanded the capabilities of natural language processing across diverse languages. Prominent models such as mBERT, XLM, and XLM-R have demonstrated remarkable zero-shot transfer learning capabilities \cite{doddapaneni2021primerpretrainedmultilinguallanguage}. These advancements have paved the way for large-scale MLLMs capable of handling multiple languages simultaneously, addressing challenges in training and evaluation \cite{qin2024multilinguallargelanguagemodel, huang2024surveylargelanguagemodels}. However, the evaluation of factual accuracy in low-resource languages, particularly Indic languages, remains largely unexplored.

A key area of related work is the assessment of factual accuracy in LLMs. \citet{shafayat2024multifactassessingfactualitymultilingual} introduced a multilingual factuality evaluation pipeline, revealing that factual accuracy is notably higher in high-resource languages like English. Similarly, \citet{fu2023largelanguagemodelsreliable} found weak correlations between LLM-generated answers and human assessments in low-resource languages, emphasizing the unreliability of factual content. FactChecker, an automatic factual error detection framework developed by \citet{wang2024earthflatunveilingfactual}, revealed factual inconsistencies in up to 45\% of model outputs across major LLMs. Additionally, \citet{wang2023surveyfactualitylargelanguage} provided a comprehensive survey on factuality challenges, highlighting key evaluation methodologies and metrics. Despite these contributions, most efforts predominantly focus on high-resource languages, with little attention given to factual accuracy in low-resource languages.


The performance disparity between high-resource and low-resource languages has also been extensively documented. Studies show that LLMs achieve significantly higher factual accuracy in languages with abundant training data, while low-resource languages often yield unreliable outputs \cite{10275753, jayakody2024performancerecentlargelanguage}. The absence of robust benchmarks for factual accuracy in low-resource languages exacerbates this disparity, preventing meaningful evaluation and improvement \cite{magueresse2020lowresourcelanguagesreviewpast}. 

Another critical challenge impacting factual accuracy in low-resource languages is hallucination — where LLMs generate factually incorrect content. Research shows that hallucinations are more prevalent in low-resource settings or when translating from high-resource languages like English \cite{li2024dawndarkempiricalstudy}. Moreover, \citet{shafayat2024multifactassessingfactualitymultilingual} highlighted the Western-centric bias in LLMs, which further affects factual accuracy in low-resource languages. While hallucination detection frameworks like HaluEval 2.0 and FActScore \cite{li2024dawndarkempiricalstudy, shafayat2024multifactassessingfactualitymultilingual} attempt to mitigate this, they remain tailored to high-resource languages, leaving low-resource languages underserved.

Low-resource languages like Indic languages face additional challenges such as data sparsity, dialectal variation, and underrepresentation of culturally relevant knowledge, further diminishing factual accuracy \cite{li2024quantifyingmultilingualperformancelarge}. Despite rapid advancements in LLMs, there is a clear void in large-scale factual accuracy evaluation for Indic languages. Our work addresses this gap by evaluating the factual accuracy of state-of-the-art LLMs across 19 Indic languages using the IndicQuest dataset. This benchmark offers the first large-scale factuality evaluation tailored to Indic languages, providing practical insights for improving factual reliability in low-resource contexts.

\section{Overview of Dataset}
This study assesses the factual accuracy of multilingual large language models (LLMs) using the IndicQuest dataset \cite{rohera2024l3cubeindicquestbenchmarkquestinganswering}. Key features of the dataset include:
\begin{itemize}
    \item \textbf{Format:} Question-answer pairs
    \item \textbf{Languages:} English and 19 Indian languages including Assamese, Bengali, Dogri, Gujarati, Hindi, Kannada, Konkani, Maithili, Malayalam, Marathi, Meitei (Manipuri), Nepali, Odia, Punjabi, Sanskrit, Sindhi, Tamil, Telugu, and Urdu
    \item \textbf{Size:} 200 pairs per language, totaling 4000 pairs
    \item \textbf{Domains:} Geography, history, politics, literature, and economics
        \item \textbf{Content:} Region-specific questions spanning India’s northern, eastern, western, and southern regions, along with some universal ones. These were manually curated in IndicQuest from sources focused on Indian regional knowledge to ensure coverage of culturally and locally relevant information.
\end{itemize}
The IndicQuest dataset plays a crucial role in our study.  Designed to test LLM knowledge in Indic languages, it addresses a critical gap in existing benchmarks by providing regionally and culturally relevant questions. The dataset was created by preparing questions in English first and then translating them into Indic languages, ensuring semantic consistency across languages. This uniformity enables more accurate evaluation of factual accuracy, as the dataset serves as a reliable ground truth for comparison during our analysis. By utilizing this resource, we can conduct a comprehensive evaluation of LLMs' performance across diverse Indian languages.

\section{Methodology} 

We utilized the IndicQuest dataset, which contains gold-standard questions and answers in 20 languages, to evaluate the performance of various multilingual large language models (LLMs). Responses were generated for these questions using selected LLMs, which were then evaluated by another LLM acting as a judge.

\subsubsection{Models}
Responses were generated for English, Hindi, and Marathi using the models Gemma-2-2b-it, Gemma-2-9b-it, Llama-3.1-8b, Llama-3.1-405b-it, and GPT-4o. For the remaining 17 Indic languages, including Assamese, Bengali, Dogri, Gujarati, Kannada, Konkani, Maithili, Malayalam, Meitei (Manipuri), Nepali, Odia, Punjabi, Sanskrit, Sindhi, Tamil, Telugu, and Urdu, the models Gemma-2-2b-it, Gemma-2-9b-it, Llama-3.1-8b, and Llama-3.1-405b-it were used, excluding GPT-4o due to resource limitations.

\subsubsection{Evaluation Metrics}
We employed two complementary sets of evaluation metrics to capture both 
subjective quality dimensions and reference-based text overlap.

\paragraph{LLM-based Evaluation Metrics:}
To evaluate the quality of model responses, we adopt a multidimensional 
scoring framework, assigning a score on a continuous scale from 1 to 5 
across five criteria:

\begin{itemize}
    \item \textbf{Factual Accuracy:} Measures the alignment between the model’s answer and provided ground truth facts. A score of 5 indicates complete agreement; 3 denotes partial correctness (approximately half the facts are correct); 1 reflects total misalignment. Intermediate scores represent varying degrees of partial accuracy.
    
    \item \textbf{Relevance:} Assesses whether the answer directly and adequately addresses the question posed. Highly relevant responses receive a 5, while irrelevant or off-topic answers receive a 1.
    
    \item \textbf{Clarity:} Captures the lucidity and coherence of the answer. Well-structured and easily understandable responses are rated 5; unclear, confusing, or poorly constructed answers are rated 1.
    
    \item \textbf{Language Consistency:} Ensures the output matches the input question’s language. Perfect consistency scores a 5; answers with mismatched language receive lower scores, with a 1 denoting complete inconsistency.
    
    \item \textbf{Conciseness:} Rewards responses that succinctly present information without unnecessary verbosity. Concise, focused answers earn a 5, while verbose, rambling responses are rated 1.
\end{itemize}

Each generated answer is scored by the LLM judge (\texttt{Llama-3.1-405b-it}) 
using a standardized evaluation prompt (see Appendix). Fractional scores 
are allowed to capture fine-grained distinctions. These dimensions are later 
averaged into an overall quality score, with greater weight on factual accuracy.

\paragraph{Automatic Reference-based Metrics:}
In addition to the LLM-judged metrics, we report reference-based automatic 
metrics to benchmark against widely used NLG evaluation standards:

\begin{itemize}
    \item \textbf{F1 Score:} Computed at the token level to balance 
    precision (correctness of generated words) and recall (coverage of gold 
    answer content).

    \item \textbf{ROUGE-L:} Measures the longest common subsequence overlap 
    between generated and gold responses, capturing both lexical overlap 
    and sequence fluency.
\end{itemize}

These metrics were calculated using the official IndicQuest gold references. 
By combining LLM-judged qualitative scoring and reference-based quantitative 
evaluation, we ensure a more reliable and multidimensional assessment of 
system performance across languages.

\subsubsection{Evaluation Process}
We adopted the evaluation methodology described in the IndicQuest work. For this purpose, we selected \texttt{Llama-3.1-405b-it} as the automated judge owing to its state-of-the-art performance on multilingual reasoning tasks and its robust support for both English and Indic languages. Recent multilingual model evaluations have demonstrated that Llama-3.1-405b-it provides superior factuality and linguistic understanding, particularly for low-resource Indic languages, making it well-suited for our diverse evaluation setting. While GPT-4o has been more widely used as a judge in the community, we opted for Llama-3.1-405b-it since it is open, reproducible, and free to use, ensuring accessibility and transparency.

The LLM judge was prompted to assess generated responses by comparing them with gold-standard answers from the dataset while considering the context of each question. It produced evaluation scores autonomously, without human intervention, with the prompt explicitly specifying the six evaluation metrics: factual accuracy, relevance, clarity, language consistency, conciseness, and an overall score. The overall score was calculated by considering all five metrics, with greater weight given to factual accuracy. The scores for each metric were assigned on a scale of 1 to 5, ensuring consistent and quantifiable evaluation across responses.  The full evaluation prompt is provided in the Appendix for reproducibility. 

We then conducted a detailed analysis of the scores obtained, focusing on domain-wise, language-wise, and model-wise performance. The analysis was primarily based on factual accuracy, but we also considered the average of all evaluation metrics to identify any significant disparities between the performance of models in English and the Indic languages.
We aimed to quantify the disparity in performance, particularly focusing on factual accuracy, which serves as a crucial metric for assessing the reliability of responses across languages and models. To further validate performance, reference-based metrics (F1 and ROUGE-L) were computed using gold answers, with results presented in the Appendix.



\section{Results}

\begin{figure}
    \centering
    \includegraphics[width= \columnwidth]{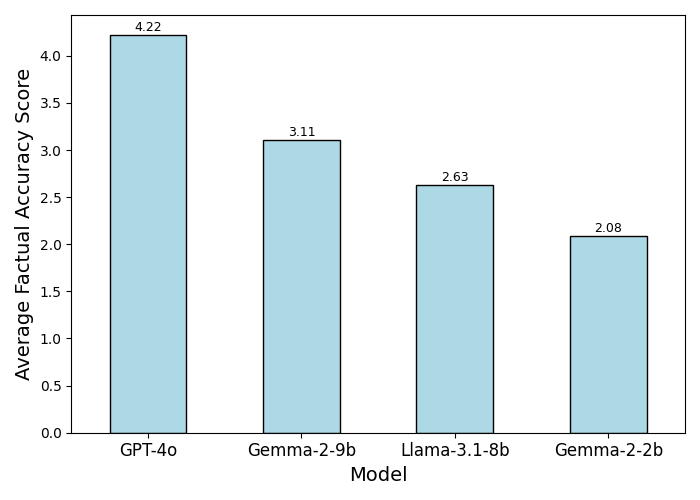}
    \caption{Average factual accuracy of responses generated by each LLM model, aggregated across all questions in the dataset.}
    \label{fig:factual_accuracy_by_model}
\end{figure}

\begin{figure}
    \centering
    \includegraphics[width= \columnwidth]{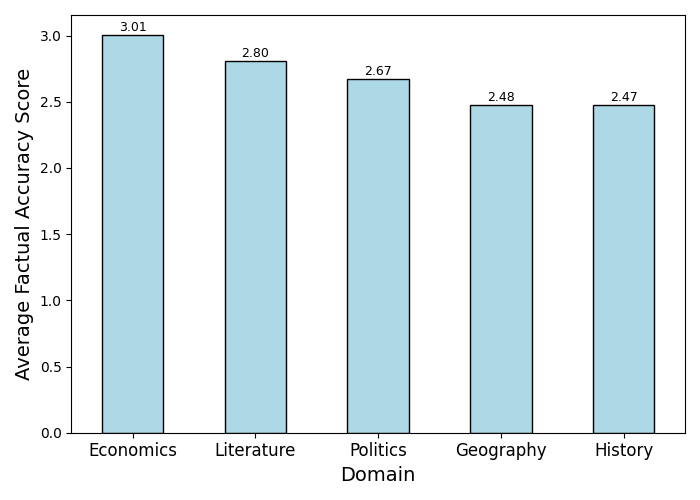}
    \caption{Average factual accuracy scores across individual domains. The scores are computed by aggregating the factual accuracy of LLM responses to all questions—across all languages—of that domain}
    \label{fig:factual_accuracy_by_domain}
\end{figure}


\begin{figure*}
    \centering
    \includegraphics[width=\textwidth]{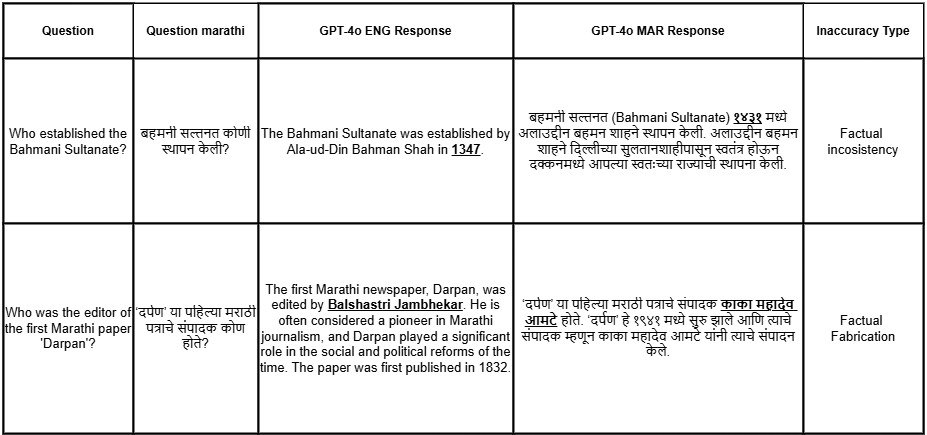}
    \caption{Factual Inaccuracy Examples}
    \label{fig:error-analysis-table}
\end{figure*}

\begin{figure}
    \centering
    \includegraphics[width= \columnwidth]{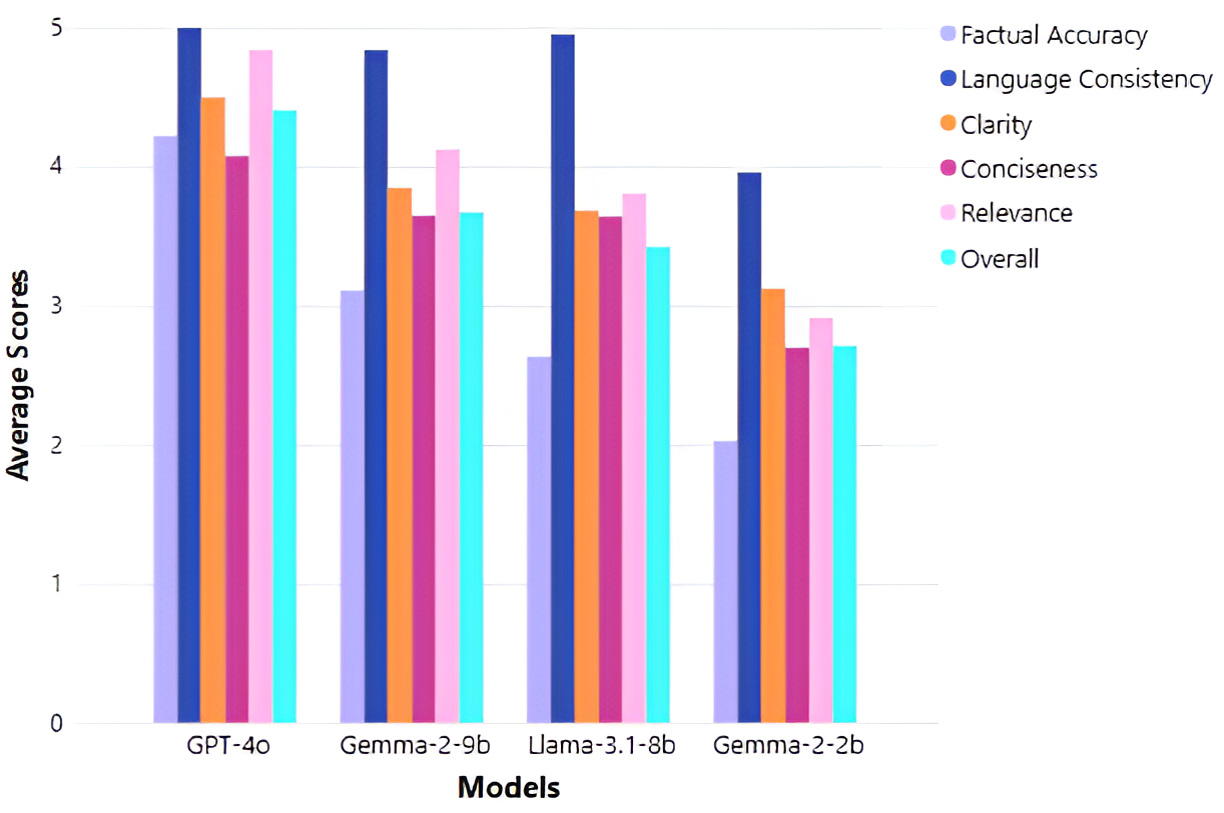}
    \caption{Model Performance Metrics}
    \label{fig:model-performance}
\end{figure}

Our analysis reveals several key findings regarding the performance of multilingual LLMs across English and Indic languages. We present these findings as follows:
\begin{enumerate}
    \item \textbf{English consistently outperforms all Indic languages across all metrics}. The performance gap is evident in Table \ref{tab:metrics_scores}, where languages like Odia and Urdu exhibit the lowest scores, highlighting the challenges faced by LLMs in low-resource languages. Based on the scores obtained, both factual accuracy (Figure \ref{fig:factual_accuracy_by_language}) and overall performance allow for the classification of Indic languages into three distinct performance-based buckets - Moderate Performance (Hindi, Marathi, Maithili, Nepali, Dogri, Sanskrit, Konkani), Low Performance (Malayalam, Telugu, Kannada, tamil, Gujarati, Punjabi, Assamese, Bengali) and Extremely Low Performance (Manipuri, Sindhi, Odia, Urdu)
    \item \textbf{Language Consistency} - 
    Models that performed well in factual accuracy typically also produced more consistent responses within a language. This suggests that factual grounding often goes hand in hand with linguistic fluency. This trend can be clearly seen in Figure \ref{fig:lc-fa-scatter-plot} in the appendix, where languages with higher factual accuracy scores are generally clustered towards higher language consistency values as well. 

    \item \textbf{Model performance} - Model ranking remains consistent across evaluation metrics (see Figure \ref{fig:factual_accuracy_by_model} and Figure \ref{fig:model-performance}), with GPT-4o outperforming all others, followed by Llama-3.1-405b-it, Gemma-2-9b, Llama-3.1-8b, and Gemma-2-2b. An exception is in Language Consistency, where Llama-3.1-8b performs slightly better than Gemma-2-9b, though still below GPT-4o. This ranking pattern aligns with model sizes, where larger models generally demonstrate superior factual accuracy, likely due to increased capacity for knowledge retention and reasoning.
    \item \textbf{Domain Influence} - The choice of domain significantly affects factual accuracy, with English consistently outperforming low-resource Indic languages across all domains. History and Geography show the lowest factual accuracy, particularly in Indic languages, likely due to their reliance on contextual interpretation and region-specific knowledge, respectively. These domains appear more challenging for models to ground responses accurately. In contrast, domains like Economics and Politics yield higher factual accuracy, possibly due to clearer structure and better representation in the training data (see Figure \ref{fig:factual_accuracy_by_domain}).
    \item \textbf{Hallucination examples} - The examples in Figure \ref{fig:error-analysis-table} demonstrate how difficult it is for LLMs to preserve the factual correctness in Indic languages with limited resources. Inaccuracies in these responses can take the form of inconsistency, where known facts are misrepresented or fabrication, where false information is generated.  These errors show how LLMs struggle with accuracy in Indic languages, despite the fact that the correct information is available in English. 
    
    We qualitatively identified examples of hallucinations—model-generated outputs containing fabricated or factually inconsistent information not supported by ground truth—and included illustrative cases in the paper. While a formal annotation protocol was not used to systematically quantify hallucinations across the entire dataset, these examples highlight common types of factual errors observed in responses. Future work will aim to develop a more rigorous and scalable hallucination detection methodology
\end{enumerate}


\section{Conclusion}
The primary goal of this study was to highlight the persistent performance gap between multilingual LLMs when responding to queries in English versus Indic languages. Our findings demonstrate that asking the same question in English consistently yields better responses, as evidenced by higher factual accuracy, overall scores, and language consistency metrics. This disparity underscores the advantages of resource-rich languages like English in current LLM training paradigms. \\
The findings highlight the need for more comprehensive quantitative analyses to assess LLMs' overall performance and factual accuracy, especially for low-resource languages. Bridging this gap requires targeted efforts in developing enriched datasets, domain-specific optimizations, and evaluation frameworks tailored to multilingual and low-resource settings. 

\paragraph{Limitations and Future Work.} 
While this study covers 200 question–answer pairs per language and spans 19 Indic languages, its limited scale constrains the generalizability of our conclusions. Expanding dataset size and domain coverage remains an important direction for future work. Likewise, GPT‑4o could not be included at scale due to paid API restrictions; instead, we provide limited comparisons to contextualize results with open models.

\newpage

\bibliographystyle{acl_natbib}
\bibliography{main.bib}

\clearpage
\appendix
\section*{Appendix}
\vspace{-1.95em}

\begin{figure}[H]
    \centering
    \includegraphics[width=\columnwidth]{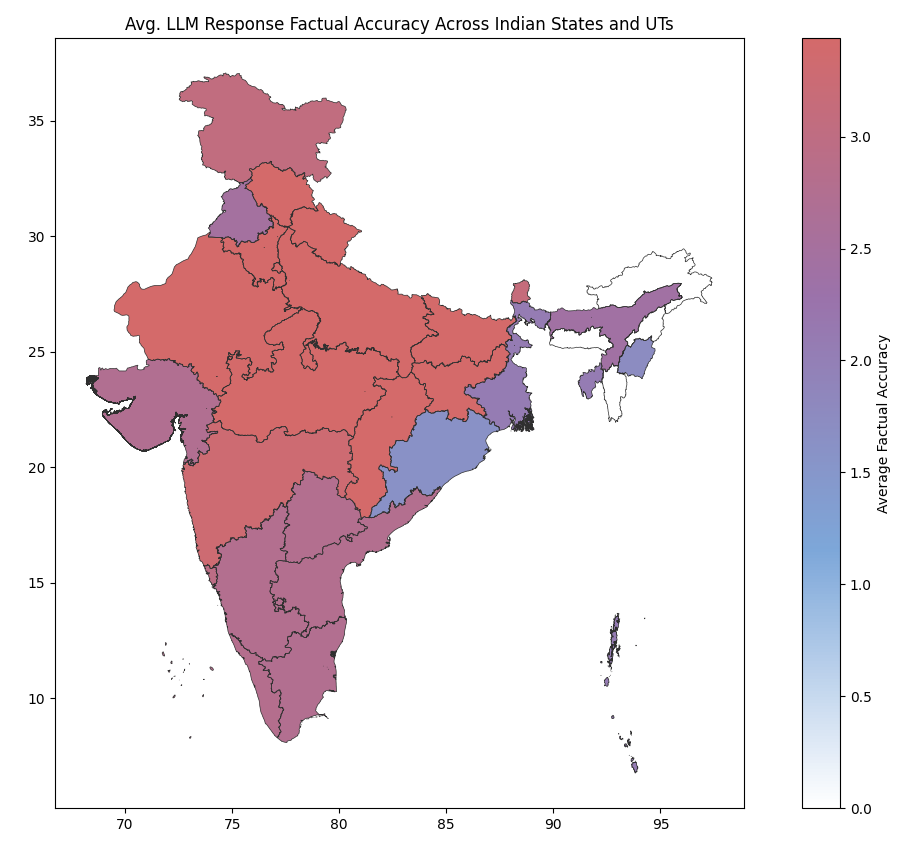}
    \caption{Heatmap illustrating the average factual accuracy of LLM-generated responses across Indian states and union territories, based on the primary language of each region included in the dataset. The average score for each region is computed from the factual accuracy of its corresponding language: Andhra Pradesh, Telangana (Telugu); Arunachal Pradesh, Meghalaya, Mizoram, Nagaland, Ladakh (excluded — no matching language in dataset); Assam (Assamese); Bihar, Chhattisgarh, Haryana, Himachal Pradesh, Jharkhand, Madhya Pradesh, Rajasthan, Uttar Pradesh, Uttarakhand, Delhi, Chandigarh (Hindi); Goa (Konkani); Gujarat, Daman \& Diu, Dadra \& Nagar Haveli (Gujarati); Karnataka (Kannada); Kerala, Lakshadweep (Malayalam); Maharashtra (Marathi); Manipur (Manipuri); Odisha (Odia); Punjab (Punjabi); Sikkim (Nepali); Tamil Nadu, Puducherry (Tamil); Tripura, West Bengal, Andaman \& Nicobar Islands (Bengali); Jammu \& Kashmir (Dogri). Regions not associated with any language in the dataset are shown in white.}
    \label{fig:factual_accuracy_heatmap_appendix}
\end{figure}

\begin{figure}[H]
    \centering
    \includegraphics[width=\columnwidth]{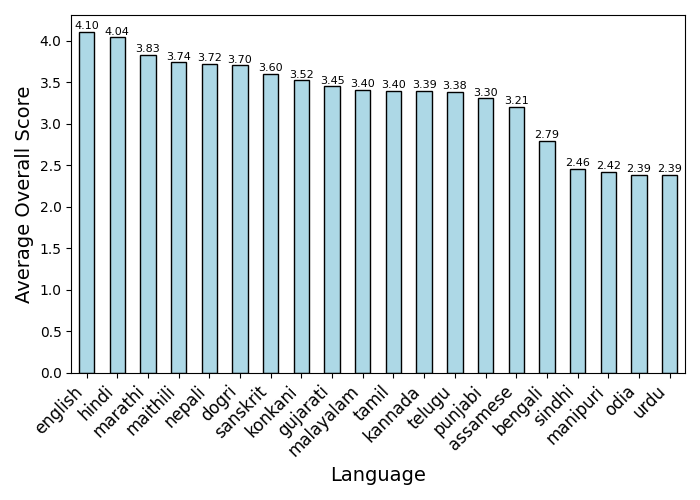}
    \caption{Overall Language Performance — Overall score assigned by the LLM judge aggregated across all responses for each language in the dataset.}
    \label{fig:language-performance-overall}
\end{figure}

\begin{figure}[H]
    \centering
    \includegraphics[width=\columnwidth]{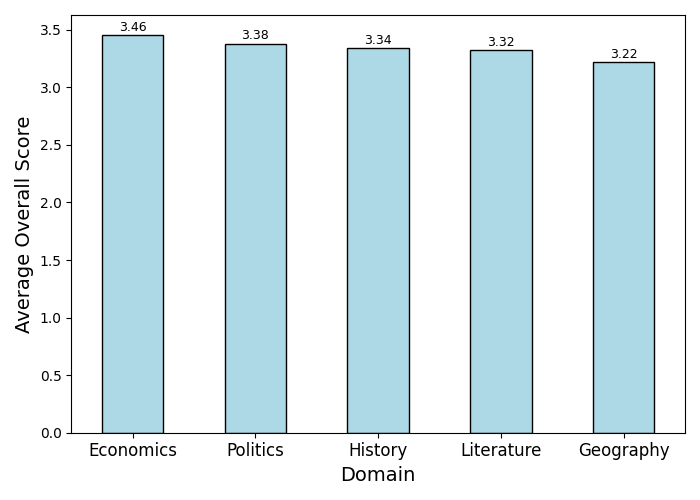}
    \caption{Overall Domain Performance — Overall score assigned by the LLM judge aggregated across all responses for each domain in the dataset.}
    \label{fig:domain-performance-overall}
\end{figure}

\begin{figure}[H]
    \centering
    \includegraphics[width=\columnwidth]{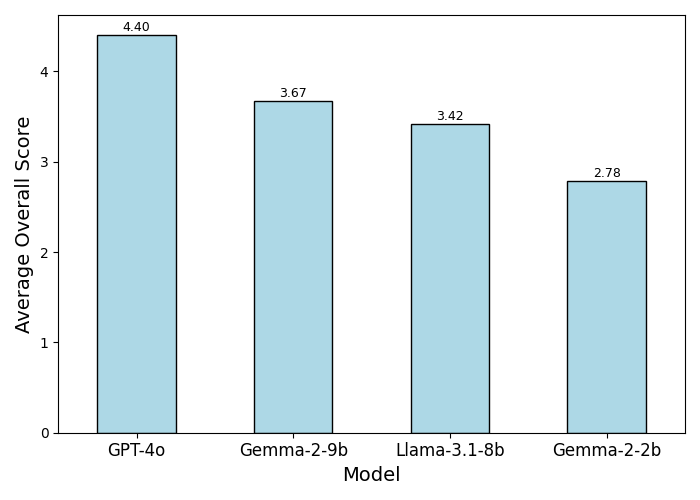}
    \caption{Overall Model Performance — Overall score given by the LLM judge aggregated across all model responses to questions in that domain from the dataset}
    \label{fig:model-performance-overall}
\end{figure}

\begin{figure}[H]
    \centering
    \includegraphics[width=\columnwidth]{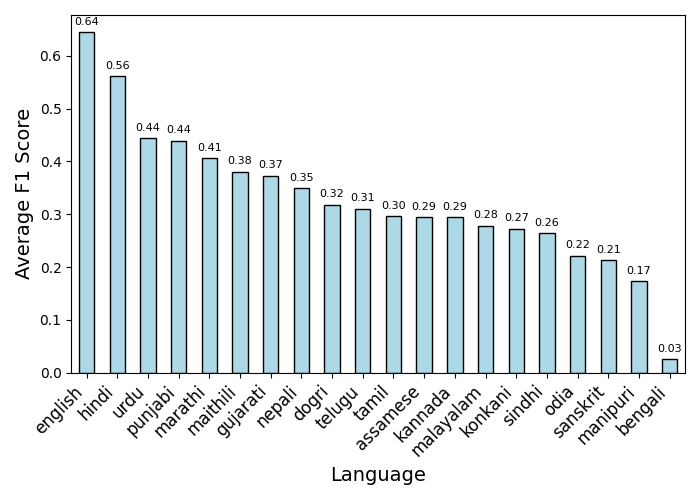}
    \caption{Language Performance on the basis of F1 Score}
    \label{fig:language-performance-f1}
\end{figure}

\begin{figure}[H]
    \centering
    \includegraphics[width=\columnwidth]{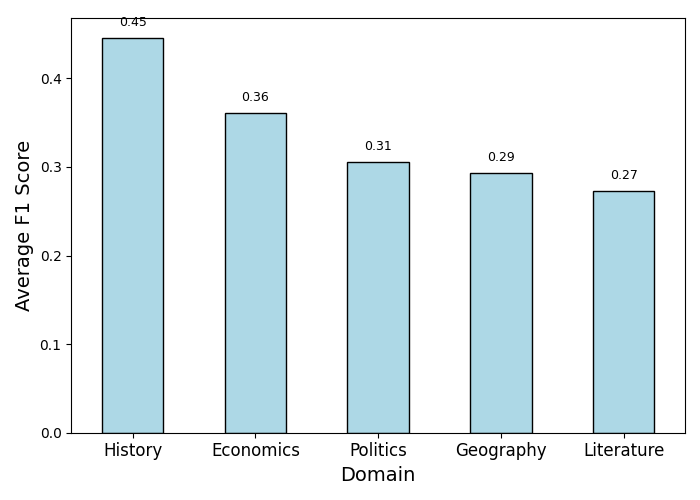}
    \caption{Domain Performance on the basis of F1 Score}
    \label{fig:domain-performance-f1}
\end{figure}

\begin{figure}[H]
    \centering
    \includegraphics[width=\columnwidth]{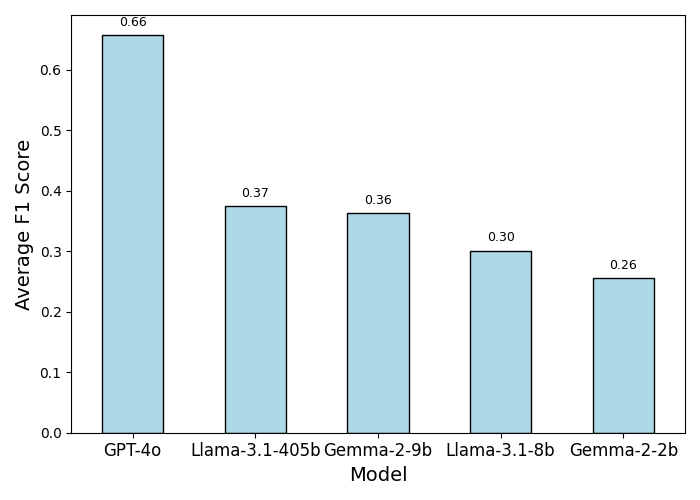}
    \caption{Model Performance on the basis of F1 Score}
    \label{fig:model-performance-f1}
\end{figure}

\begin{figure}[H]
    \centering
    \includegraphics[width=\columnwidth]{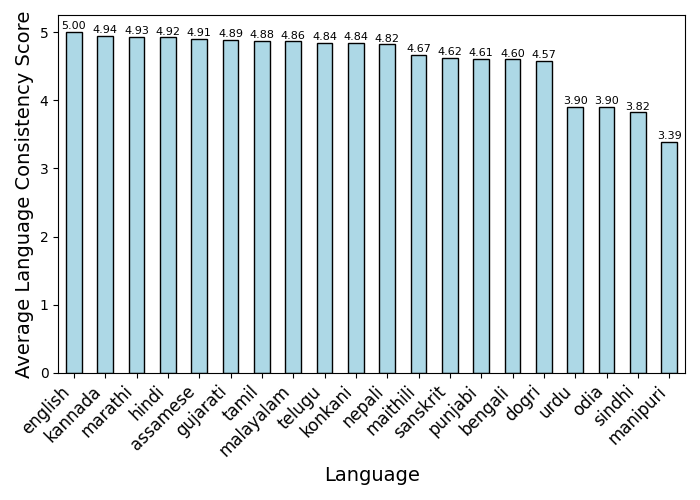}
    \caption{Language Performance on the basis of average Language Consistency Score (assigned by LLM Judge)}
    \label{fig:language-performance-language-consistency}
\end{figure}

\begin{figure}[H]
    \centering
    \includegraphics[width=\columnwidth]{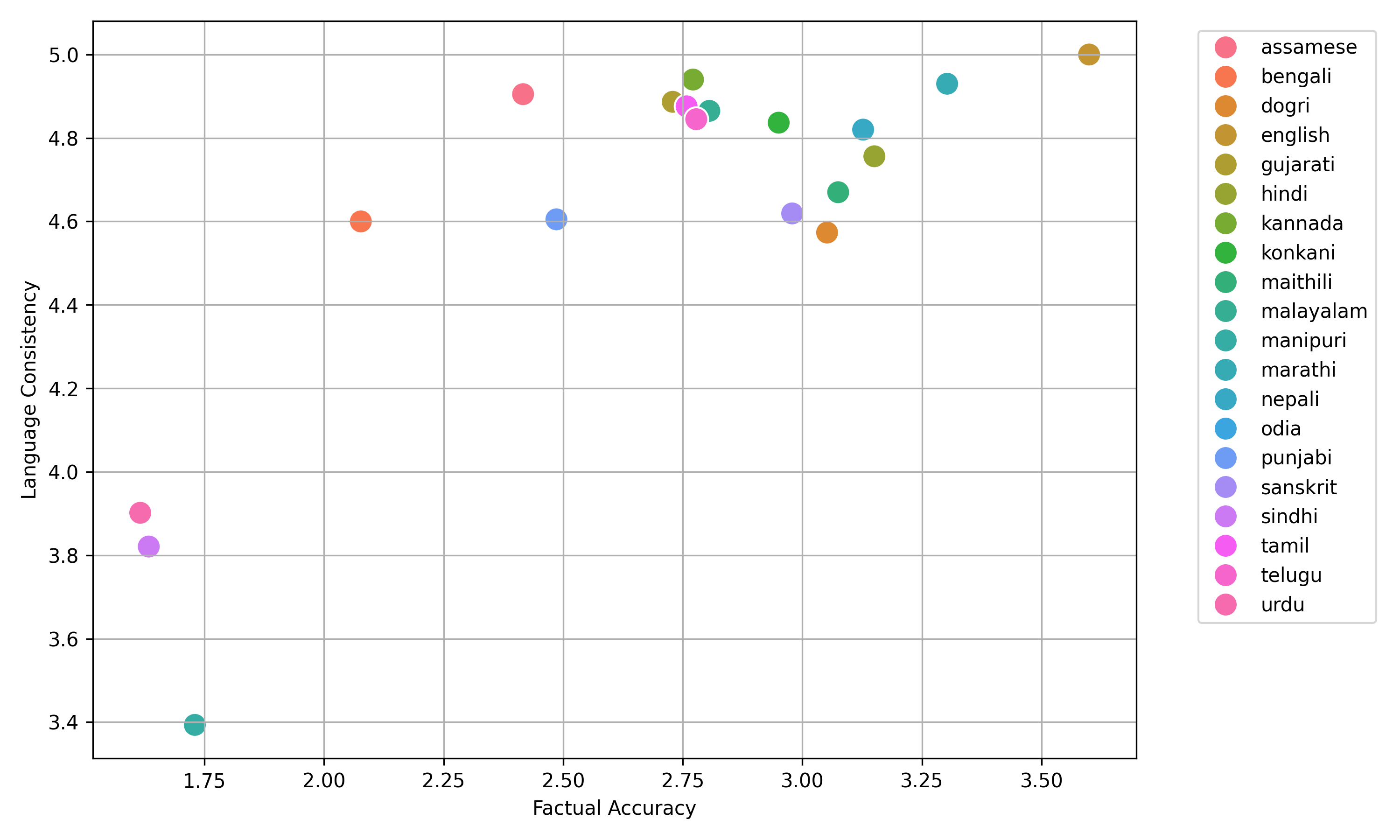}
    \caption{Relationship between Factual Accuracy and Language Consistency across 20 Languages}
    \label{fig:lc-fa-scatter-plot}
\end{figure}

\lstset{
  language=Python,
  basicstyle=\tiny\ttfamily, 
  keywordstyle=\color{blue},
  commentstyle=\color{black},
  stringstyle=\color{red},
  numbers=none,
  frame=single,
  breaklines=true,
  rulecolor=\color{black},
  linewidth=\columnwidth, 
  xleftmargin=0.05\columnwidth, 
  xrightmargin=0.05\columnwidth 
}

\begin{lstlisting} [caption={Evaluaton prompt given to Llama-3.1-405b-it}, label={lst:prompt-code}]
prompt = f"""
Evaluate the quality of the model's responses to questions from a benchmark dataset on a scale of 1-5 (score can be a decimal fraction format number) across the following parameters:

Factual Accuracy: Given an input question, ground truth facts relevant to the question, and the model/bot's answer, evaluate how well the information in the model's answer aligns with the provided ground truth facts. Assign a score on a scale of 1 to 5 based on the following criteria: a score of 5 indicates complete alignment with all ground truth facts; a score of 3 represents partial alignment where approximately half of the facts are correct; and a score of 1 denotes complete misalignment with the ground truth facts. Scores between these benchmarks can reflect varying degrees of alignment or discrepancies.

Relevance: Assess how well the model's answer directly addresses the question. A score of 5 indicates a highly relevant answer, while a score of 1 indicates an irrelevant or off-topic response.

Clarity: Evaluate the clarity and coherence of the model's answer. A score of 5 means the answer is well-structured and easy to understand, while a score of 1 means it is confusing or poorly constructed.

Language Consistency: Ensure that the language of the response matches the language of the question unless otherwise specified. Penalize cases where there is a mismatch between the input language specified in the question and the response language.

Conciseness: Rate how concise the answer is while still providing necessary information. A score of 5 indicates the answer is succinct and to the point, while a score of 1 indicates excessive verbosity or unnecessary information.

Input Details:
Question: {question}
Ground Truth Facts: {ground_truth}
Model/Bot Answer: {model_answer}
After evaluating each parameter, provide an overall rating on a scale of 1-5 considering all the parameters. The parameter factual accuracy should have more weightage in the overall score.

Output Format:
Return the evaluation scores in the following JSON format(Return only the JSON and nothing else):
  {{
    "Factual Accuracy": score,
    "Relevance": score,
    "Clarity": score,
    "Language Consistency": score,
    "Conciseness": score,
    "Overall": average_score
  }}
"""
\end{lstlisting}

\end{document}